\documentclass[runningheads]{llncs}
\usepackage{graphicx}
\usepackage{amsmath,amssymb} 
\usepackage{color}
\usepackage[boxed,commentsnumbered]{algorithm2e}
\usepackage[breaklinks=true,bookmarks=false, colorlinks, citecolor=green]{hyperref}

\begin{document}

\newcommand{\point}{
    \raise0.7ex\hbox{.}
    }


\pagestyle{headings}

\mainmatter

\title{Learning Action Concept Trees and Semantic Alignment Networks from Image-Description Data} 

\titlerunning{Learning Action Concept Trees from Image-Description Data} 

\authorrunning{Jiyang Gao, Ram Nevatia} 

\author{Jiyang Gao, Ram Nevatia} 
\institute{University of Southern California} 

\maketitle

\begin{abstract}
Action classification in still images has been a popular research topic in computer vision. Labelling large scale datasets for action classification requires tremendous manual work, which is hard to scale up. Besides, the action categories in such datasets are pre-defined and vocabularies are fixed. However humans may describe the same action with different phrases, which leads to the difficulty of vocabulary expansion for traditional fully-supervised methods. We observe that large amounts of images with sentence descriptions are readily available on the Internet. The sentence descriptions can be regarded as weak labels for the images, which contain rich information and could be used to learn flexible expressions of action categories.  We propose a method to learn an Action Concept Tree (ACT) and an Action Semantic Alignment (ASA) model for classification from image-description data via a two-stage learning process. A new dataset for the task of \emph{learning actions from descriptions} is built. Experimental results show that our method outperforms several baseline methods significantly.
\end{abstract}

\section{Introduction}
Action classification in still images has been a popular research topic in computer vision. Traditional fully-supervised learning methods for action classification rely on  large amount of fully-labelled data(\emph{i.e.} each image is labelled with one or more action categories) to learn action classifiers.  However, labelling image data with action categories requires tremendous manual work, which is time-consuming and hard to scale-up.  Another drawback of traditional supervised learning framework is that the action categories are pre-defined and limited, while humans may describe the same action with different phrases, for example, take out the chopping board and fetch out the wooden board. This drawback leads to the difficulty of vocabulary expansion, as CNN \cite{krizhevsky2012imagenet} \cite{simonyan2014very} models or SVM classifiers just assign a label to the test image. Hence, CNN or SVM models would fail to classify the categories that are not in the training set.

We observe that large amounts of images with sentence descriptions are readily available on the Internet, such as videos with captions and social media, such as Flickr and Instagram. Such sentence descriptions can be regarded as weak labels of the images. Sentence descriptions are generated by humans and contain rich information about actions, which could be used to learn an expanding vocabulary of actions. Some example are shown in Fig. 1. Another observation we make is that action concepts are naturally represented as a hierarchy; for example, ``play guitar" and ``play violin" are subcategories of ``play instrument". If such hierarchical structure of action categories is available, classification methods can choose to use detailed knowledge if necessary or generalized knowledge when details are unavailable or irrelevant.

\begin{figure}
\centering
\includegraphics[scale=0.5]{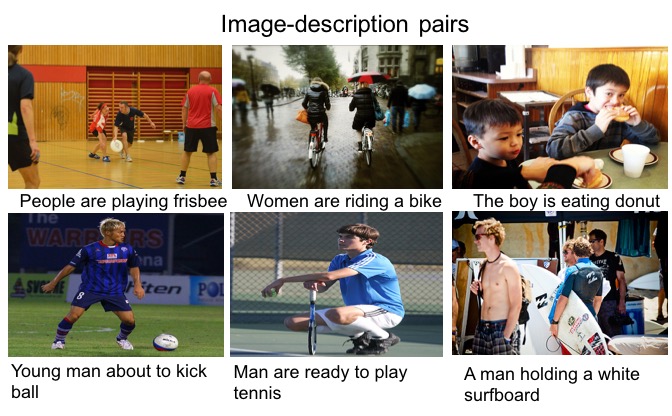}
\caption{Descriptions, as weak labels to images, contain rich information about actions. Images and corresponding descriptions are from Visual Genome \cite{krishnavisualgenome}}
\end{figure}

In this paper, we propose a method to tackle the problem of \emph{learning actions from descriptions}: Given a set of image-description data (assuming descriptions containing human action information), learning to recognize human actions. Our method supports hierarchical clustering of action concepts and vocabulary expansion for action classification. Specifically, our method learns an Action Concept Tree (ACT) and an Action Semantic Alignment (ASA) model for classification via a two-stage learning process. ASA model contains a CNN to extract image-level features, an LSTM to extract text embeddings and a multi-layer neural network to align these two modalities. In the first stage, (a) we design a Hierarchical Action Concept Discovery (H-ACD) method to automatically discover action concepts from image-description data and cluster them into a hierarchical structure (\emph{i.e.} ACT); (b)  ASA is initialized by the image-description mapping task in stage-1. In the second stage, the target action categories are matched to the nodes in ACT and the associated image data are used to fine-tune ASA for this action classification task to improve the performance. Note that no image data from test domain are used for training. 

To facilitate research on this task, we constructed a dataset based on Visual Genome \cite{krishnavisualgenome}, called Visual Genome Action (VGA). Although Visual Genome contains well-annotated region descriptions, we do not use the region information and treat the descriptions as image-level. There are 52931 image-description pairs in the training set and 4689 images of 45 categories in test set. More details of this dataset are given in Section 4.1 later.

In summary, our main contributions are:

(1) A Hierarchical Action Concept Discovery (H-ACD) algorithm to automatically discover an Action Concept Tree (ACT) from image-description data and gather samples for each action node in ACT. 

(2) An end-to-end CNN-LSTM Action Semantic Alignment (ASA) network which aligns semantic and visual representation to classify actions with expanding vocabulary.

(3) A dataset for the problem of \emph{learning actions from descriptions}, which is built on Visual Genome, containing 52931 image-description pairs for training and 45 action categories for testing.

The paper is organized as follows. Section 2 discusses the related works. In section 3, we will introduce our two-stage framework to learn actions from image-description data. We evaluate our model in section 4 and give our conclusions in section 5.

\section{Related Work}

\textbf{Action Classification in Still Images:}
The use of convolutional neural network (CNN) has brought huge improvement in action classification \cite{guo2014survey}. \cite{oquab2014learning} finetunes the CNN pre-trained on ImageNet and shows improvement over traditional methods. \cite{gkioxari2014r} designs a multi-task (person-detection, pose-estimation and action classification) model based on R-CNN.  \cite{gkioxari2015contextual} develops an end-to-end deep convolutional neural network that utilizes contextual information of actions. HICO \cite{Chao_2015_ICCV}  introduces a new benchmark for recognizing human-object interactions, which contains a diverse set of interactions with common object categories, such as ``hold banana" and ``eat pizza".  Ramanathan \emph{et al.} \cite{ramanathan2015learning} proposes a neural network framework to jointly extract the relationship between actions and uses them for training better action retrieval models. These methods all rely on fully-labelled data.

\textbf{Weakly Supervised Action Concept Learning}
Weakly supervised action concept learning relies on weakly-labelled data, such as video-caption stream data \cite{rohrbach15cvpr} \cite{torabi2015using} and focuses on automatically discovering and learning action concepts. \cite{alayrac2015learning} designs a method to automatically discover the main steps for specific tasks, such as ``make coffee" and ``change tire", from narrated instructional videos. Their method solves two clustering problems, one in text and one in video, applied one after each other and linked by joint constraints to obtain a single coherent sequence of steps in both modalities. Ramanathan \emph{et al.} \cite{Ramanathan_2013_ICCV} propose a method to learn action and role recognition models based on natural language descriptions of the training videos. Yu \emph{et al.} \cite{yu2014instructional} discover Verb-Object (VO) pairs from the captions of the instructional videos and use the associated video clips as training samples. The learned classifiers are evaluated in event classification, compared with well defined action categories in HMDB51 \cite{kuehne2011hmdb} and UCF50 \cite{reddy2013recognizing}. \cite{sun2015automatic} proposes a general concept discovery method from image-sentence corpora and apply the concepts on image-sentence retrieval tasks.

ACD \cite{gao2016acd} solves a similar problem to ours. It automatically discovers action concepts from image-sentence corpora \cite{lin2014microsoft} \cite{young2014image}, clusters them and trains classifier for each action concept cluster. However, there are two main drawbacks in this method: (1) no hierarchical clustering: once the action concepts are clustered, the detailed information are lost; (2) no vocabulary expansion: if the target test action categories are missed in the training set, ACD would fail to perform classification. 

\textbf{Language \& Vision:}
Image captioning methods take an input image and generate a text description of the image content. Recently, methods based on convolutional neural networks and recurrent neural networks \cite{donahue2015long} \cite{vinyals2015show} have shown to be an effective way on this task. VSA \cite{karpathy2015deep} is one of the recent successful models. It uses bidirectional recurrent neural networks over sentences, convolutional neural networks over image regions and a structured objective that aligns the two modalities through a multimodal embedding. Besides image captioning, other relevant work includes natural language object retrieval \cite{hu2015natural} or segmentation \cite{hu2016segmentation}, which takes an input image and a query description and outputs a corresponding object bounding box or a segmentation mask.
\section{Actions from Descriptions}
In this section, we introduce the learning framework, which is a two-stage method. In the first stage, our target is to learn a general knowledge base of actions, which contains two parts: a hierarchical structure for action concepts and a general visual-semantic alignment model. In the second stage, the framework learns to classify specific action categories (\emph{i.e.} target categories for test). The classifiers are fine-tuned from the visual-semantic alignment model learned in stage 1. The overall system is shown in Fig. 1. 

\subsection{Stage 1: Learning General Action Knowledge}
As for general action knowledge, we refers to two concepts. The first one is a hierarchical structure of actions, which we call Action Concept Tree (ACT): each node in ACT contains an action concept, such as play frisbee and play basketball, and the related images; the action concepts are extracted from descriptions and the images come from the original image-description dataset. The second one is a general visual-semantic alignment model: the input of the model is an image and a description, and the output is a confidence score of the similarity of the image and the description.

\begin{figure}
\centering
\includegraphics[scale=0.4]{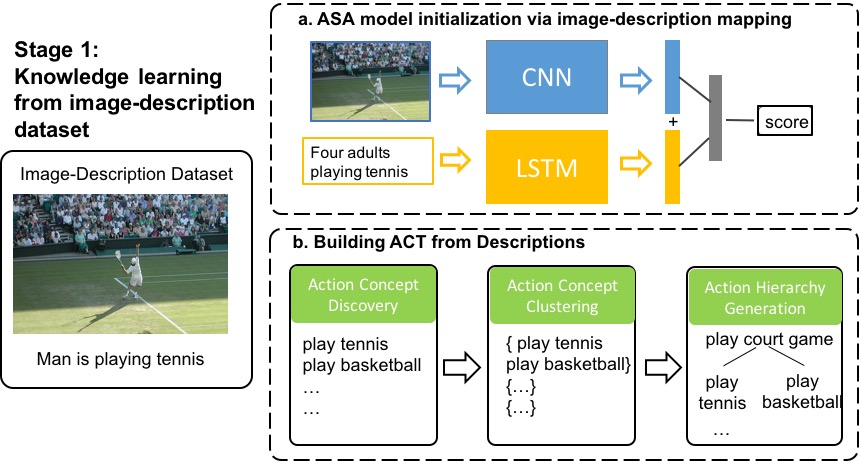}
\caption{Stage-1: Model Initialization via Image-description matching and hierarchy action discovery}
\end{figure}

\subsubsection{Hierarchical Action Concept Discovery (H-ACD)}
ACD \cite{gao2016acd} proposed an action concept discovery method working with image-sentence corpora. However, the discovered action clusters are not organized in a hierarchical structure, which may lose the detailed information after clustering. Hence, based on ACD, we propose a Hierarchical Action Concept Discovery (H-ACD) method, which automatically discovers action concepts from image-description data and organizes them in a hierarchical structure using WordNet \cite{miller1995wordnet}. The process of action concept discovery and clustering are similar to ACD. First we extract Verb-Object (VO) pairs from sentence descriptions and the visualness of these VO pairs are verified by two fold cross-validation. After visualness verification, we generate a multi-modal representation for each action concept and calculate similarity score for each pair of action concepts.

After computing the similarity, we use the H-ACD algorithm to generate a hierarchical structure for action concepts. Note that nearest neighbor (NN) clustering algorithm is proposed in ACD \cite{gao2016acd}; we use it as a part of our H-ACD algorithm. We first apply NN-clustering algorithm (we fix the parameter C of NN-clustering as 4.) \cite{gao2016acd} on all the action concepts to get a list of action clusters. Then, inside each cluster, we continuously apply NN-clustering algorithm to get more smaller clusters; we do this recursively util no new cluster is generated. Each cluster is regarded as a node in the hierarchical structure and the node names are generated following a similar naming strategy of HAN \cite{cao2016abstract} described in the following. For the object part, we find the lowest common hypernym in WordNet. For the verb part, we follow a simple strategy: if the verbs are the same, then the father node keeps the same verb; if the verbs are different, the father node is named as ``interact with". For example, for a node containing \{hold dish, hold pan\}, the least shared parent of dish and pan is container and for the verb part, ``hold" itself is the least shared parent. So the name of this action node is ``hold container". The H-ACD algorithm is shown in Algorithm 1.

\begin{algorithm}
  \SetAlgoLined
  \KwData{Concept similarity matrix $M$ of size $l \times l$ and concept list $L$ of size $l$}
  \KwResult{Action Concept Tree (ACT)}
  Queue q $\leftarrow$ NN-Clustering($M$, $L$)\;
  TreeNode root\;
  root.addChild(q.all())\;
  \While {\text{q} not empty}{
  	$L_{cluster}$ $\leftarrow$ q.pop()\;
	node=ACT.getNode($L_{cluster}$)\;
  	$M_{cluster} \leftarrow$ getSimMat($L_{cluster}$)\; 
	tinyclusters=NN-Clustering($M_{cluster}$, $L_{cluster}$)\;
	q.push(tinyclusters)\;
	node.addChild(tinyclusters)\;
   }
   Generate node names following the naming strategy. 
  \caption{Hierarchical Action Concept Discovery (H-ACD) algorithm}
\end{algorithm}

\subsubsection{ASA Model Initialization via Image-Description Mapping}

Our final target is to classify action categories. Rather than training classifiers for each category, we want to build a connection between the semantic meaning and visual meaning of actions. Therefore, we formulate the action classification as a visual-semantic alignment problem between the image and action categories.  The Action Semantic Alignment (ASA) model contains three parts: a CNN network to extract feature vector of the input image, an LSTM network to extract text embedding and an alignment network to compute the alignment score of the visual and semantic representations. Image-description mapping serves as a parameter initialization method for ASA model, which helps the model to learn a connection between semantic and visual spaces.

The input of ASA is an image $I_i$ and the corresponding sentence description $D_i$. The image is processed by VGG-16, which outputs a $d_{img}$ dimensional feature $v_i$. For a text sequence $S = (w1, ..., wT )$ with $T$ words, each word is transformed to a $d_{w2v}$ dimensional vector by the word embedding matrix and then processed by an LSTM module sequentially. The word embedding matrix is trained by skip-gram model on English Wiki data. At the final time step $t=T$, LSTM outputs the final hidden state and we use it as the sentence-level embedding $s_i$, which is a $d_{text}$ dimensional vector. $v_i$ and $s_i$ are concatenated to $vs_{ii}$ with a length of $d_{vs}=d_{text}+d_{img}$, which is visual-semantic representation of the image and description. Then we train a two-layer alignment network, with a $d_{alg}$ dimensional hidden layer. The alignment network take $d_{vs}$ dimensional input and output a confidence score $cs_{ii}$, which indicates whether the image and the description is aligned. The alignment network is implemented in a fully convolutional way as two $1*1$ convolutional layers (with ReLU function between them). 

During the training time, we optimize the model inside each mini-batch. The loss function is as follows.

\begin{equation}
\begin{split}
loss_{1}=\sum_{i=0}^{N} [\alpha_{c}\text{log}(1+\text{exp}(-cs_{i,i}))+ 
 \sum_{j=0,j\neq i}^{N}\alpha_{w}\text{log}(1+\text{exp}(cs_{i,j}))]
\end{split}
\end{equation}

where $N$ is the batch size, $\alpha_{c}$ and $\alpha_{w}$ are the loss weights for correct and wrong image-description. The loss function encourages the network to output high score of correct image-description pairs and low score of incorrect image-description pairs. In practice, we find that training converges faster using higher loss weights for correct pairs and we use  $\alpha_{c}=1$ and $\alpha_{w}=0.01$.

\begin{figure}
\centering
\includegraphics[scale=0.4]{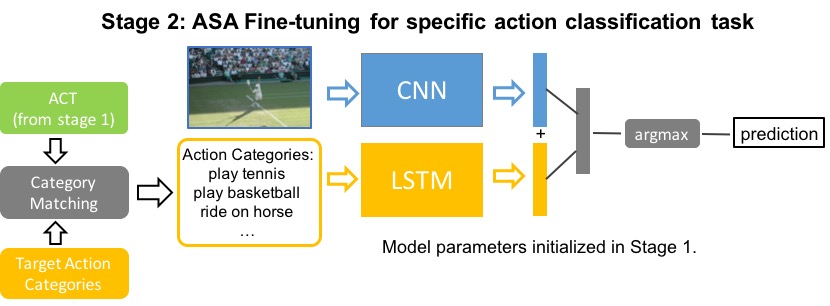}
\caption{ Stage-2: Adjust the model to a specific action classification task.}
\end{figure}

\subsection{Stage 2: Action Classification on Target Categories}
Given a set of action categories for classification (without training samples), we adjust the ASA model to the specific action classification task. The first step is to match the given action categories to some existing action nodes in ACT. Then we use the matched action nodes and the associated images to fine-tune our ASA model.

\subsubsection{Target Action Categories Matching}
We first match the actions via keyword searching. Suppose the target action category is $c_i$ and the action node in ACT is represented by $n_j$. We extract the verb and object from the target action category $c_i$ and search for them in the discovered action hierarchy to see if there is an exact match. For example, a target action category is ``play instrument" and there is a node in action hierarchy named "play instrument", then we match them and use the similarity score (calculated by ASA, see below) between them as a baseline score $\theta$. If there is no exact match via keyword searching, we assign $\theta$ with a constant value. In the second step, we use the ASA model to compute a similarity score between the target action category $c_i$ and all action nodes $n_j$ in ACT. The associated images of $n_j$ are $P_j=\{I_{jk}\}$, which has size $m$. The similarity score between $c_i$ and $n_j$ is 

\begin{equation}
S(c_i,n_j)=\frac{1}{m} \sum_{k=0}^{m}\text{ASA}(I_{jk},c_i)
\end{equation}
For a specific action category $c_i$, we select the action node $n_j$ that has the highest similarity score and if the score is larger than or equal to $\theta$, we match $<c_i, n_j>$. Note that some categories may still not be matched after the second step. After matching, we obtain a list of training samples $\{c_i,P_j\}$. The labels are $c_i$ and the training images are the associated images of the corresponding matched node $n_j$.  We don't assign any training data for the target categories with no matched node in ACT. The matching algorithm is detailed in Algorithm 2 below.

\begin{algorithm}
  \SetAlgoLined
  \KwData{ASA model, ACT and Target action categories $C=\{c_i\}$}
  \KwResult{Matched pairs $<c_i,n_j>$}
  \For {$c_i$ in $C$}{
  	\For {$n_j$ in ACT}{
		if $c_i.name$=$n_j.name$: 
			match $<c_i,n_j>$\;
		break\;
		}
	if $c_i$ is matched:\\
	\ \ \ \ ExactMatch, $\theta$ $\leftarrow$ $n_j, S(c_i,n_j)$\;
	\ \ \ \ match $<c_i,\text{ExactMatch}>$\;
	else:\\
	\ \ \ \ ExactMatch, $\theta$=None, InitializationValue\;
	MaxScore=0\;
	\For {$n_j$ in ACT and $n_j \neq $ \text{ExactMatch}}{
		if $S(c_i,n_j) > $MaxScore: \\
			\ \ \ \ Node, MaxScore $\leftarrow$ $n_j, S(c_i,n_j)$
	}
	if MaxScore $>=\theta$:\ match $<c_i,\text{Node}>$\;

   }
  \caption{Target action categories matching algorithm}
\end{algorithm}

\subsubsection{ASA Fine-tuning for Specific Action Classification Task}

We use the training samples obtained in last step to fine-tune the network. In stage 1, the loss function tends to match the correct image-description pair and it works as a parameter initialization method. In stage 2, our goal is to optimize the model to some specific classification task. We formulate the classification problem as a image-description matching problem. The name of the category is regarded as a text sequence, just like the sentence description. Suppose there are $M$ categories, leading to $M$ corresponding category descriptions $\{\text{CD}_j,j=0,1,2...,M-1\}$. Suppose the label of the input image $I_i$ is $t_i$, then the loss function of stage 2 is as follows.
\begin{equation}
\begin{split}
loss_{2}=\sum_{i=0}^{N} [\alpha_{c}\text{log}(1+\text{exp}(-cs_{i,t_i}))+ 
\sum_{j=0,j\neq t_i}^{M}\alpha_{w}\text{log}(1+\text{exp}(cs_{i,j}))]
\end{split}
\end{equation}

where $cs_{i,j}$ is the matching score between $I_i$ and $\text{CD}_j$, $N$ is the batch size. We use $\alpha_c=1$ and $\alpha_{w}=0.01$. The loss function encourages the correct image-action pairs to output high positive score and other wrong pairs output low negative score.

\subsubsection{Action Category Prediction} At test time, the prediction of an input image $I_i$ is the argmax of the matching scores $cs_{i,j}$ between $I_i$ and $\text{CD}_j$. 

\begin{equation}
\text{prediction}(I_i)=\text{argmax}(cs_{i,j}), \ j=0,1,2...M-1
\end{equation}

\section{Evaluation}
\subsection{Experiments on VGA}
\subsubsection{Dataset: Visual Genome Action (VGA).}
There are many image-description datasets, which are suitable for \emph{learning actions from descriptions}. However, none of them contain pre-defined action categories and category annotations for each image. Therefore, we construct a dataset from Visual Genome for this problem, called Visual Genome Action (VGA). We split Visual Genome into two parts: 75\% for training and validation and 25\% for testing. The training set and test set are carefully checked to ensure that there is no overlap of images between these two sets. 

For the training split, since we only focus on human action learning, we filter out the descriptions which don't have verbs or human subjects; for example, ``a dog is running on the grass" and ``a man with a white shirt" are filtered out. 52931 image-description pairs remain after such filtering. The descriptions in Visual Genome are region based, but we treat them as image-level descriptions. For the test split, we extract Verb-Object (VO) pairs and filter out the ones with very few image samples. After that, we manually filter out the VOs with no visual meaning, such as ``do things". Finally, there are 45 categories and 4689 images for testing. The 45 test action categories are listed in Table 1. Some categories overlap; for example, ``hold racket" and ``play tennis", ``hit ball" and ``play soccer". We manually checked each image of these categories and added additional labels if necessary.  For example, if an image of the category ``hold racket" also represents the action of ``play tennis", then we also add this image to the category of ``play tennis".  In other cases, people may be just holding a tennis racket but not playing, then we don't add additional labels to such images.

\setlength{\tabcolsep}{6pt}
\begin{table}
\centering
\caption{Action categories in VGA test set}
\begin{tabular}{| c c c c c |}

\hline
boat&brush tooth &color hair &do trick &drink wine \\
eat fruit& eat pizza &enjoy outdoors& fly kite& hit ball\\
hold bag& hold banana& hold bat& hold camera &hold controller\\
 hold dog& hold fork& hold kite& hold knife& hold pole\\
 hold racket & hold sandwich& hold umbrella& jump& play baseball \\
 play basketball&play frisbee&play soccer & play tennis&read book\\
 ride elephant&ride horse&ride wave&run&sit\\
ride skateboard&ski&smile&stand&surf\\
swim&use phone&walk&watch game &wear necklace\\
\hline\end{tabular}

\end{table}

\subsubsection{Metric.}
We tested our model on action classification task on VGA. As for evaluation metric, we report the mean Average Precision (mAP), Recall@1 and Recall@5.

\subsubsection{Network implementation.} We implemented ASA network in Tensorflow\cite{tensorflow2015-whitepaper}, including CNN network, LSTM network and the multi-layer alignment network. For the CNN part, We use VGG-16 architecture and the parameters are initialized by ImageNet\cite{deng2009imagenet} image classification dataset. We use a standard LSTM architecture with 1000-dimensional hidden state. The descriptions input to LSTM have maximum length of 6 for both stage-1 and stage-2. The hidden layer of the visual-semantic alignment network is 500-dimensional. We train a skip-gram \cite{mikolov2013distributed} model for the word embedding matrix using the English Dump of Wikipedia. The dimension of the word vector is 500. The whole network is trained end-to-end in two stages. We use three Adam optimizers \cite{kingma2014adam} to optimize CNN, LSTM and the visual-semantic alignment network. The learning rates are 0.0001, 0.001 and 0.001 respectively. The model is trained on a Tesla K40 GPU; the batch size is 96. It takes about 1 day to train the whole model for both stage-1 and stage-2.

\subsubsection{System variants.} We experimented with variants of our system to test the effectiveness of our method. \textbf{ASA (Stage 1):} we only trained the ASA model for stage-1 using the image-description pairs. \textbf{ASA (Stage 2):} we only trained the ASA model for stage-2 using the matched action nodes in ACT.  \textbf{ASA (Stage 1+2, w/o ACT):} ASA model is trained for stage-1 and stage-2, but we only use flat action concepts (\emph{i.e.} only the leaf action nodes in ACT) to match the target action categories.  \textbf{ASA (Stage 1+2, w/ ACT):} this is our full model; ASA model is trained for stage-1 and stage-2, and full ACT is used to match the target action categories.

\subsubsection{Baseline methods.}
We introduce the baseline methods we implemented.

ACD \cite{gao2016acd}+SVM+AdaBoost: In this baseline method, we use ACD\cite{gao2016acd} to discover a list of action concepts from the training set and train SVM classifiers \cite{fan2008liblinear} for each action concept. Then we match the test action categories with the discovered action concepts by keyword searching. Multiple action concepts may matched to the same test categories and each of them can be regarded as weak classifier to the test category. To make use of all the related training data, we further use AdaBoost to build a stronger classifier.

ACD \cite{gao2016acd}+DeViSE \cite{frome2013devise}: In this baseline method, we first use ACD to discover a list of action concepts. Instead of training SVM classifiers for each of them, we apply DeViSE\cite{frome2013devise} methodology. The verb and the object of a action category are transformed to vectors using a word embedding matrix and are concatenated together. The word embedding matrix is trained by wiki dump data and the dimension of the word vector is 500. All the discovered action concepts and the associated images are used to train DeViSE model. At test time, the action categories are transformed to vectors using the same word embedding matrix. The prediction of an input image is the argmax of the matching scores between the image and the test categories.

Visual-semantic alignment \cite{karpathy2015deep}: This baseline method is similar to the model in VSA \cite{karpathy2015deep}. However, we use regular LSTM instead of BRNN to encode the input description. The image is processed by VGG-16 and 4096 dimensional fc7 vector is extracted as image-level feature.  The training data are image-description pairs. The output of the model is a confidence score which indicates whether the image and description are matched. The test image is matched with all action categories and the prediction is the category with the highest score.

\subsubsection{Action Concept Tree (ACT).}

\begin{figure}
\centering
\includegraphics[scale=0.38]{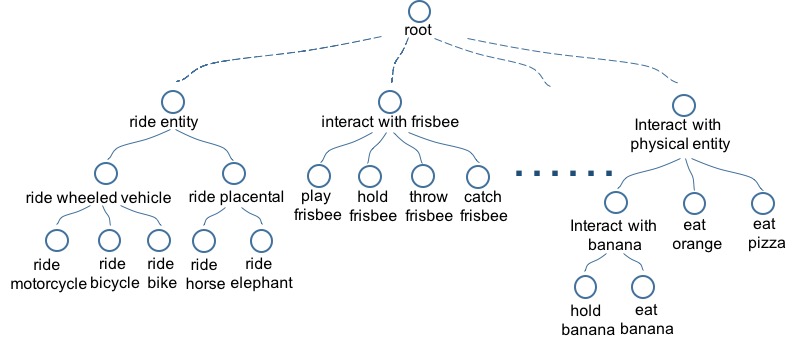}
\caption{Example nodes in ACT.}
\end{figure}

There are totally over 100 action concepts (\emph{i.e.} leaf action nodes) discovered in the training set of VGA. These action concepts are clustered into a 4-layer action concept tree (ACT). Due to the limited space, we can't illustrate the whole ACT. Some nodes in ACT are shown in Fig. 4. Under the node ``ride entity", we can see that ``ride motorcycle", ``ride bicycle" and ``ride bike" are clustered together and the automatically generated node name is ``ride wheeled vehicle", as ``wheeled vehicle" is the lowest common hypernym of ``bike", ``bicycle" and ``motorcycle" in WordNet.  Under the node ``interact with frisbee", there are four leaf nodes: ``catch frisbee", ``hold frisbee", ``play frisbee" and ``throw frisbee". They have common object "frisbee" but different verb actions, so we generate the father node name as ``interact with frisbee".  The node ``interact with physical entity" is illustrated as a poor case of our naming strategy. The child nodes have no common verb and the lowest common hypernym of the objects in WordNet is ``physical entity", therefore the father node is named as ``interact with physical object", which is a very vague action name. Although, the naming strategy is not ideal in this case, the cluster itself still represents one meaningful action category: ``interact with food".

\setlength{\tabcolsep}{6pt}
\begin{table}
\centering
\caption{Comparison of different methods on the VGA action classification test set}
\begin{tabular}{l| l l l}

\hline
Method &mAP(\%)&R@1(\%)&R@5(\%)\\ \hline
ACD+SVM+AdaBoost & 20.2&24.5&56.3\\
ACD+DeViSE & 22.1 &25.1&54.2\\ 
VSA &15.9&18.1&47.3\\ \hline
ASA (Stage 1) &20.1&25.3&56.5\\
ASA (Stage 2) &18.5&24.6&50.4\\
ASA (Stage 1+2, w/o ACT) &26.8&29.6&60.4\\
\textbf{ASA (Stage 1+2, w/ ACT)} &\textbf{28.5}&\textbf{31.3}&\textbf{63.2}\\
\hline\end{tabular}

\end{table}

\subsubsection{Action Classification Results.}

The experimental results on VGA are shown in Table 2. From the results, we can see that our 2-stage learning method outperforms several baseline methods.  Training models with only stage-1 or stage-2 would lower the performance. Stage-1 only learns general image-description matching knowledge and it does not optimize the model to a specific action classification task; on the other hand, without stage-1, stage-2 optimizes the model from random parameters and it may overfit on such a small dataset of language. Using the hierarchical structure of action concepts ( \emph{i.e.} ASA(Stage1+2, w/ ACT))brings a 1.7\% improvement, compared with the flat structure of action concepts ( \emph{i.e.} ASA(Stage1+2, w/o ACT)). We believe the reason is that ACT and the node matching algorithm together provide a better way to organize and search for the generalized and detailed knowledge of actions. For example, compared with the flat action concept structure, the test category ``brush tooth" is matched not only with the node of ``brush tooth", but also with the parent node of ``hold toothbrush" and ``brush tooth" in ACT, which allows ASA to use the additional data provided by``hold toothbrush".



In Fig. 5, some example predictions are shown. 
We can see that failure could happen when subtle human-object interaction differences are involved; for example, ``hold sandwich" and ``hold banana" have the same verb action (\emph{i.e.} hold) and visually similar objects. 

\subsection{Experiments on Flickr30k and PASCAL VOC}
We use the same experiment setup as ACD \cite{gao2016acd}: using Flickr30k \cite{young2014image} as source image-description dataset and PASCAL VOC 2012 action classification as target test dataset. Flickr30k contains 30000 images and each image is captioned by 5 sentences. PASCAL VOC 2012 action classification dataset has 10 action categories. We train our full model (ASA + ACT) on Flickr30k and apply the action concepts on PASCAL VOC.

\setlength{\tabcolsep}{3pt}
\begin{table}\small
\centering
\caption{Transfer learning from Flickr30k to PASCAL VOC 2012 action classification test set (AP\%).  }

\begin{tabular}{|l|l l l l l l l l l l|r|}
\hline
method  &jump&phone&instr.& read &bike &horse&run& photo& comp.& walk &mAP\\ \hline
ACD\cite{gao2016acd}   &62.2&15.4 &78.8 &\textbf{29.6} & 84.5 & 85.9& \textbf{60.8}& 24.0& 69.2& \textbf{32.4}& 54.3\\ \hline
ASA+ACT & \textbf{63.5} &\textbf{15.5} &\textbf{80.9} &28.9 & \textbf{86.7} &\textbf{92.0}& 60.7 & \textbf{24.1}& \textbf{69.3}& 30.9& \textbf{55.2}\\ \hline

\end{tabular}
\end{table}

As shown in Table 3, our method outperforms ACD \cite{gao2016acd} in most categories and by 0.9\% in mAP. For example, "ride bike" and "ride horse" are two separate subcategories in our ACT and provide more precise data for training, while ACD \cite{gao2016acd} may cluster these two with other categories such as "ride skateboard". 

\begin{figure}
\centering
\includegraphics[scale=0.62]{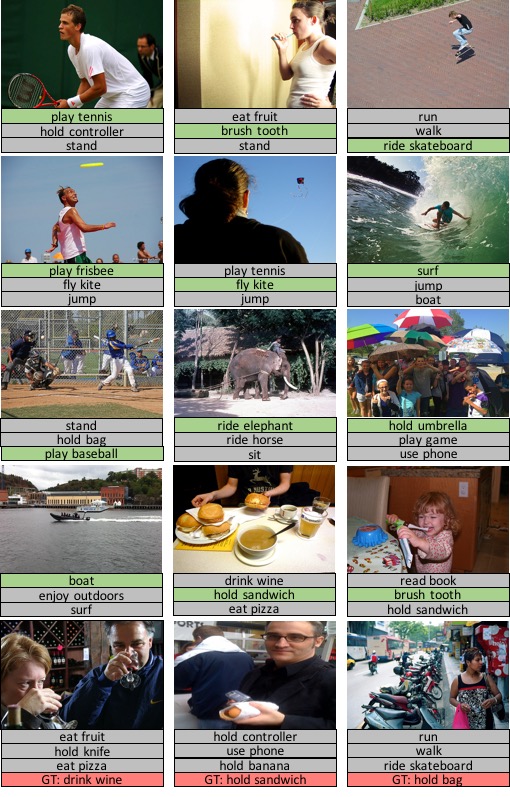}
\caption{ Prediction examples of the top 3 results on VSA test set. The first four rows are positive examples and green represents the condition when the prediction matches the ground truth. The last row shows some failure cases.}
\end{figure}

\section{Conclusion}
We presented a two-stage learning framework to learn an Action Concept Tree (ACT) and an Action Semantic Alignment (ASA) model from image-description data. Stage-1 has two steps: (a) ACT is discovered and built by H-ACD algorithm, each node in the tree contains an action name and the relevant images; (b) ASA model is trained by image-description mapping task for parameter initialization. In stage two, we adjust the ASA model to a specific action classification task. The first step is to match the target action categories to the action nodes in ACT discovered in stage-1. After matching, we use the associated data to fine-tune ASA model to this action classification task. Experimental results show that our model outperforms several baseline methods significantly. 

\vspace{3mm}
\noindent {\bf Acknowledgement}. This research was supported, in part, by the Office of Naval Research under grant N00014-13-1-0493. We would like to thank Chen Sun for valuable discussions.

\bibliographystyle{splncs}
\bibliography{egbib}

\end{document}